\documentclass{article}

 \usepackage[main, final]{neurips_2025_workshop}
\workshoptitle{Foundation Models for the Brain and Body}

\usepackage[utf8]{inputenc} 
\usepackage[T1]{fontenc}    
\usepackage{hyperref}       
\usepackage{url}            
\usepackage{booktabs}       
\usepackage{amsfonts}       
\usepackage{nicefrac}       
\usepackage{microtype}      
\usepackage{xcolor}         
\usepackage{todonotes}      
\usepackage{geometry}
\usepackage{makecell}
\usepackage{longtable}
\usepackage{tabularx}
\usepackage{pdflscape}
\usepackage{adjustbox}
\usepackage{changepage}
\usepackage{subfig}
\usepackage{pgfplots}
\usepackage{pgfplotstable}
\usepackage{amssymb}
\usepackage{amsmath}
\usepackage{multirow} 
\usepackage{rotating} 
\usepackage{graphicx} 
\usepackage{colortbl}  
\usepackage{tikz}      
\usepackage{array}
\usepackage{graphicx,calc}
\newlength\myheight
\newlength\mydepth
\settototalheight\myheight{Xygp}
\newlength{\iconheight}

\newcommand{\texticon}[1]{%
    \settoheight{\iconheight}{\textbf{X}}
    \includegraphics[height=1.5\iconheight]{#1}
}

\newcommand{\hs}[1]{\hspace*{#1cm}}

\newcolumntype{R}[1]{>{\raggedleft\arraybackslash}m{#1}}
\newcolumntype{C}[1]{>{\centering\arraybackslash}m{#1}}
\title{EEG-Bench: A Benchmark for EEG Foundation Models in Clinical Applications}

\author{%
  Ard Kastrati\\
  ETH Zurich\\
  \texttt{akastrati@ethz.ch} \\
  \And
  Josua B\"{u}rki \\
  ETH Zurich \\
   \texttt{jbuerki@ethz.ch} \\
   \AND
  Jonas Lauer \\
  ETH Zurich \\
   \texttt{jlauer@ethz.ch} \\
  \And
  Cheng Xuan \\
  ETH Zurich \\
   \texttt{cxuan@ethz.ch} \\
  \And
  Raffaele Iaquinto \\
  ETH Zurich \\
   \texttt{iaquintr@ethz.ch} \\
  \And
  Roger Wattenhofer \\
  ETH Zurich \\
   \texttt{wattenhofer@ethz.ch} \\
}

\begin{document}
\maketitle

\begin{abstract}
We introduce a unified benchmarking framework focused on evaluating EEG-based foundation models in clinical applications. The benchmark spans 11 well-defined diagnostic tasks across 14 publicly available EEG datasets, including epilepsy, schizophrenia, Parkinson’s disease, OCD, and mild traumatic brain injury. It features minimal preprocessing, standardized evaluation protocols, and enables side-by-side comparisons of classical baselines and modern foundation models. Our results show that while foundation models achieve strong performance in certain settings, simpler models often remain competitive, particularly under clinical distribution shifts. To facilitate reproducibility and adoption, we release all prepared data and code in an accessible and extensible format.

\end{abstract}

\section{Introduction}

 The rise of foundation models has started to revolutionize healthcare, with large language models (LLMs) achieving impressive results in medical reasoning, and clinical summarization~\cite{medpalm2, healthbench}. While language remains the primary focus, recent work is increasingly exploring multimodal foundation models that extend to vision, audio, and biosignals. Electroencephalography (EEG) is an essential modality in clinical neurology, widely used to diagnose and monitor brain disorders such as epilepsy, sleep disorders, etct. It is non-invasive, cost-effective, and offers high temporal resolution, making it a valuable tool across diverse clinical environments. However, EEG signals are notoriously difficult to work with: they are noisy, low in spatial resolution, and highly variable across subjects and hardware setups. To address these issues, recent work has proposed EEG-specific foundation models adapted from advances in language~\cite{wei2022emergent}, vision~\cite{awais2023foundational}, and audio processing~\cite{huang2023masked}. Models such as BENDR~\cite{bendr},
 Neuro-GPT~\cite{cui2024neurogpt}, and LaBraM~\cite{jiang2024large} leverage large-scale public EEG datasets and self-supervised learning objectives to build general-purpose neural representations that transfer across subjects, tasks, and datasets. For evaluation of EEG foundation models, they are often fine-tuned on tasks
 from the brain-computer interface domain,
 as well as on clinical datasets like TUEG Epilepsy \cite{tuep}, TUAB Abnormal EEG detection \cite{TUHEEG}, and sleep staging. However, EEG is also used to support diagnosis of many other conditions, including mild traumatic brain injury (mTBI), Parkinson’s disease (PD), schizophrenia, and obsessive-compulsive disorder (OCD), which remain underrepresented in current evaluation efforts.

\begin{figure}
    \centering
    \begin{minipage}[t][][b]{.32\textwidth}
        \resizebox{\textwidth}{!}{\begin{tikzpicture}
        \begin{axis}[
            xbar,
            symbolic y coords={
                Unknown, Easycap, actiCAP, NeuroScan SynAmps2, Brain Vision\\system
            },
            width=4cm,
            height=5cm,
            ytick=data,
            y tick label style={text width=1.8cm, align=right, font=\footnotesize\linespread{0.8}\selectfont},
            xlabel={\makecell{Number of Datasets per\\Hardware System used}},
            bar width=10pt,
            xmin=0,
            xmax=7.5,
            nodes near coords,
        ]
        \addplot coordinates {
            (5,Unknown)
            (1,Easycap)
            (1,actiCAP)
            (1,NeuroScan SynAmps2)
            (6,Brain Vision\\system)
        };
        \end{axis}
        \end{tikzpicture}}
    \end{minipage}
    \hfill
    \begin{minipage}[t][][b]{.67\textwidth}
        \includegraphics[width=\textwidth, trim={0 1cm 0 .5cm}]{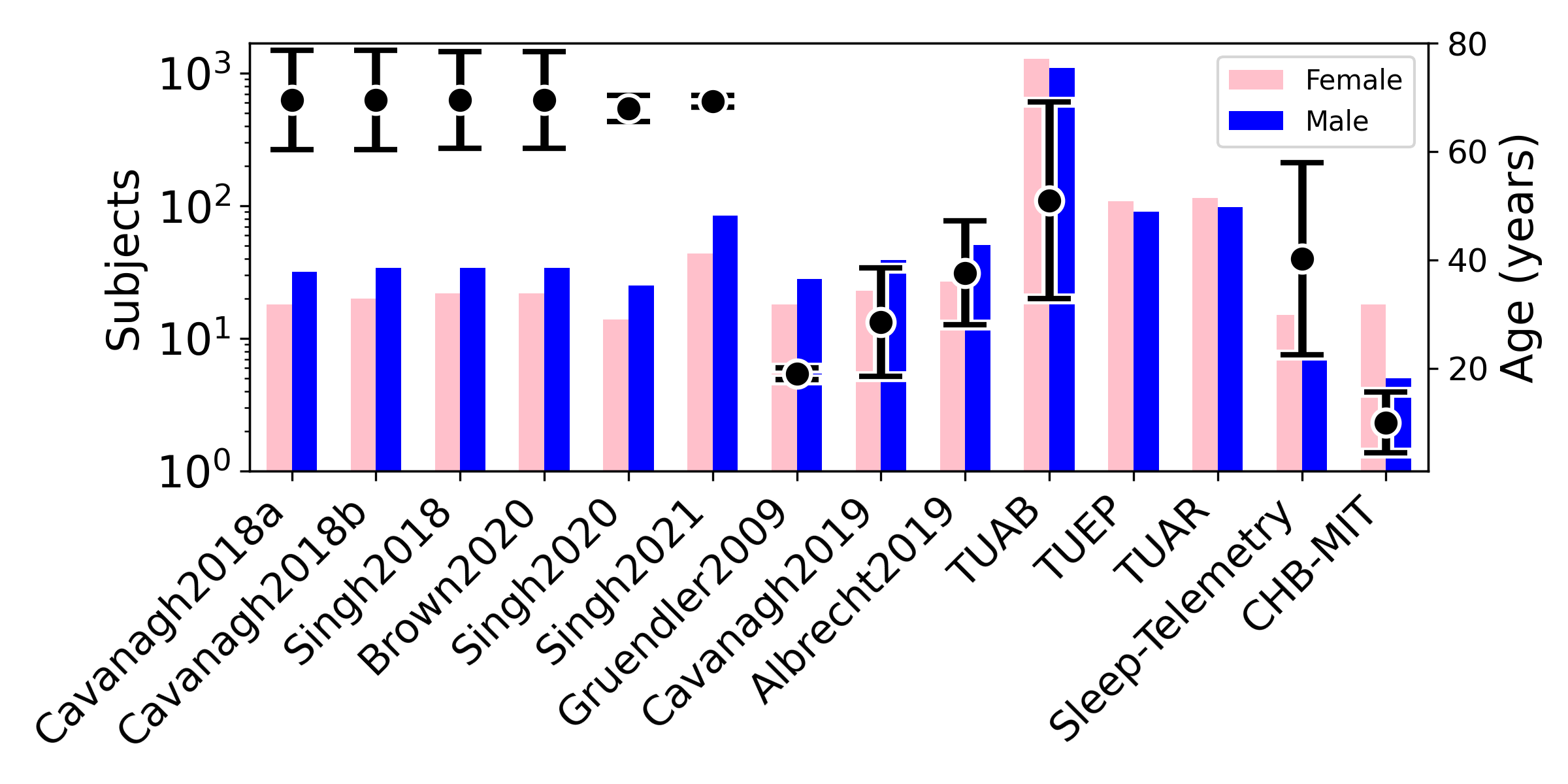}
    \end{minipage}


    \caption{
Dataset diversity across three dimensions: (Left) Distribution of EEG hardware systems used across datasets, each with potentially different electrode layouts, channel counts, and sampling rates. (Right) Gender and age distribution of subjects, covering a broad range from infants (1 year old) to elderly adults (up to 80 years), reflecting the inclusion of both pediatric and geriatric populations. The age bars show mean and standard deviation. 
}\label{fig:diversity}
\end{figure}
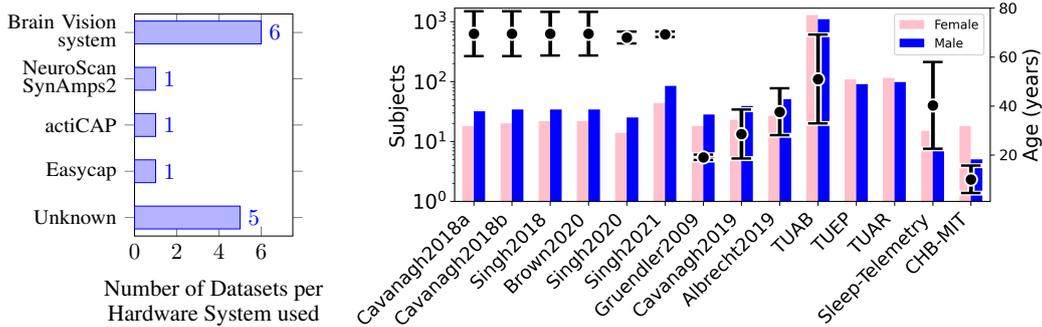

To address this, we present a unified benchmarking framework to evaluate EEG models across a wide range of clinical tasks and datasets. It includes 14 publicly available datasets spanning diverse clinical conditions, subject demographics, recording paradigms (e.g., sleep, rest), and hardware setups (e.g., different EEG caps and sampling rates), enabling evaluation under realistic and heterogeneous conditions (Figure~\ref{fig:diversity}, Table~\ref{tab:clinical_datasets_table}). From this data, we define 11 diagnostic tasks of varying difficulty (Table~\ref{tab:benchmarking_tasks}). We then perform the first comprehensive comparison of EEG foundation models and classical baselines in the clinical setting. Our benchmark provides a consistent, extensible platform for evaluating generalization and highlights the current strengths and limitations of general-purpose EEG models. All datasets and code are released in a reproducible, plug-and-play format to accelerate research in clinical EEG modeling and support the development of robust and generalizable neural decoding systems.

\section{Benchmark Framework}
To systematically evaluate the generalization capabilities of EEG decoding models, we introduce a benchmarking framework that integrates diverse datasets, defines a standardized set of decoding tasks and evaluation metrics, and supports evaluation across model types and domains.

\begin{table}[t]
    \centering
    \setlength\tabcolsep{1.5pt}
\begin{tabularx}{\textwidth}{|>{\raggedright\arraybackslash\columncolor[gray]{0.9}}p{2.2cm}|C{1.4cm}|>{\raggedright\arraybackslash}X|C{0.4cm}|C{0.9cm}|C{0.8cm}|C{0.8cm}|}
    \toprule
    \textbf{Dataset} & \textbf{Task} & \textbf{Experimental Paradigm} & \rotatebox{90}{\textbf{Channels}\hs{.5}} & \rotatebox{90}{\textbf{Total Length}} & \rotatebox{90}{\textbf{Subjects}\hs{.6}} & \rotatebox{90}{\textbf{Recordings}\hs{.2}}\\
    \midrule
    
    Cavanagh2018a \cite{CAVANAGH2018} & PD &
    \multirow{3}{*}{
\begin{tikzpicture}[baseline=(current bounding box.center), scale=0.8, every node/.style={scale=0.8}]
\draw[|-|, thick] (0,0) -- (6.8,0);
\foreach \x/\xtext in {6.8/200ms} {
    \draw (\x, 0.1) -- (\x, -0.1) node[below] {\xtext};
}
\node[align=center, above] at (0.8, 0) {Oddball:};
\node[align=center, above] at (4.1, 0) {\includegraphics[width=5.5cm]{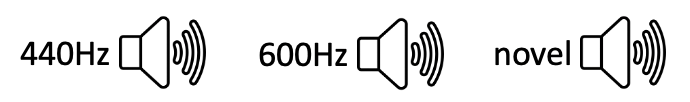}};
\node[align=center, below] at (4, 0) {x100 trials, x2 blocks};
\draw[|-|, thick] (7,0) -- (7.8,0);
\foreach \x/\xtext in {7.8/2m} {
    \draw (\x, 0.1) -- (\x, -0.1) node[below] {\xtext};
}
\node[align=center, above] at (7.5, 0.1) {\includegraphics[width=0.7cm]{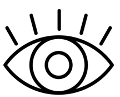}};
\end{tikzpicture}
}
    & 64 & 21 h & 50 & 77 \\

    Cavanagh2018b \cite{CAVANAGH2018} & PD &

    & 64 & 4 h & 56 & 56 \\
    Cavanagh2019 \cite{Cavanagh2019mtbi} & mTBI &
    & 64 & 57 h & 62 & 62 \\

    Albrecht2019 \cite{ALBRECHT2019131} & \makecell[c]{Schizo-\\phrenia} &
    \begin{tikzpicture}[baseline=(current bounding box.center), scale=0.8, every node/.style={scale=0.8}]
    \draw[|-|, thick] (0,0) -- (7.8,0);
    \foreach \x/\xtext in {7.8/850ms} {
        \draw (\x, 0.1) -- (\x, -0.1) node[below] {\xtext};
    }
    \node[align=center, above] at (1, 0) {Simon task:};
    \node[align=center, above] at (4.8, 0.1) {\includegraphics[width=6cm]{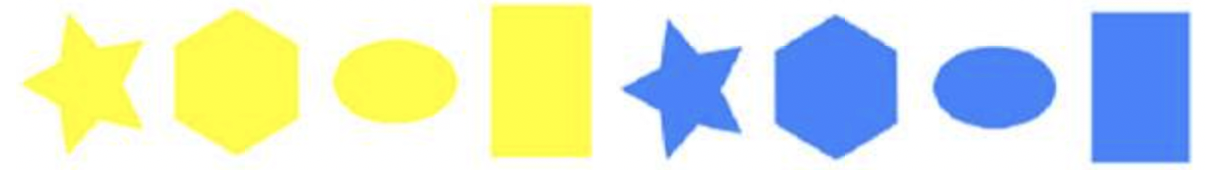}};
    \node[align=center, below] at (4, 0) {> x80 trials, x4 blocks};
    \end{tikzpicture}
    & 64 & 51 h & 78 & 78 \\

    Singh2018 \cite{Singh2018} &  PD &
    \begin{tikzpicture}[baseline=(current bounding box.center), scale=0.8, every node/.style={scale=0.8}]
    \draw[|-|, thick] (0,0) -- (7.8,0);
    \foreach \x/\xtext in {7.5/750ms-1.5s} {
        \draw (\x, 0.0) -- (\x, -0.0) node[below] {\xtext};
    }
    \node[align=center, above] at (1, 0) {Simon task:};
    \node[align=center, above] at (4.8, 0.1) {\includegraphics[width=6cm]{images/simon_task.png}};
    \node[align=center, below] at (4, 0) {> x20 stimulus, x4 blocks};
    \end{tikzpicture}
    & 64 & 108 h & 56 & 56 \\

    Brown2020 \cite{Brown2020} &  PD &
    \begin{tikzpicture}[baseline=(current bounding box.center), scale=0.8, every node/.style={scale=0.8}]
    \draw[|-|, thick] (0,0) -- (7.8,0);
    \foreach \x/\xtext in {7.5/4s} {
        \draw (\x, 0.0) -- (\x, -0.0) node[below] {\xtext};
    }
    \node[align=center, above] at (1, 0) {Reward\\learning:};
    \node[align=center, above] at (4.8, 0.1) {\includegraphics[width=6cm]{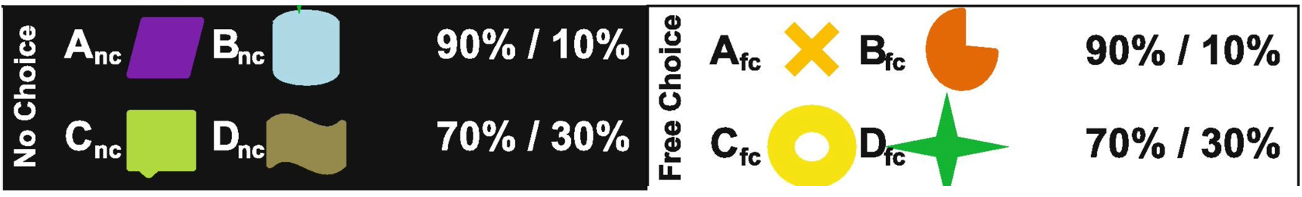}};
    \node[align=center, below] at (4, 0) {x20 stimulus, x3-x5 blocks};
    \end{tikzpicture}
    & 64 & 31 h & 56 & 84 \\

    Gruendler2009 \cite{Gruendler2009} & OCD &
    \begin{tikzpicture}[baseline=(current bounding box.center), scale=0.8, every node/.style={scale=0.8}]
    \draw[|-|, thick] (0,0) -- (1.9,0);
    \draw[|-|, thick] (2,0) -- (4.9,0);
    \draw[|-|, thick] (5,0) -- (7.8,0);
    \foreach \x/\xtext in {4.9/4s, 7.8/4s} {
        \draw (\x, 0.0) -- (\x, -0.0) node[below] {\xtext};
    }
    \node[align=center, above] at (1, 0) {Reward\\learning:};
    \node[align=center, above] at (3.5, 0.1) {\includegraphics[width=3cm]{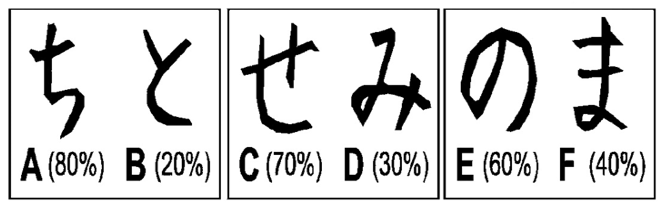}};
    \node[align=center, above] at (6.5, 0.1) {\includegraphics[width=3cm]{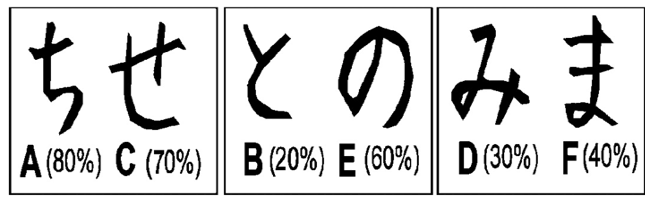}};
    \node[align=center, below] at (3.3, 0) {x60 trials, x6 blocks};
    \node[align=center, below] at (6.5, 0) {testing phase};
    \end{tikzpicture}
    & 64 & 22 h & 46 & 46 \\

    Singh2020 \cite{SINGH2020} &  PD &
    \begin{tikzpicture}[baseline=(current bounding box.center), scale=0.8, every node/.style={scale=0.8}]
    \draw[|-|, thick] (0,0) -- (8,0);
    \foreach \x/\xtext in {2/0.5s, 4/1-2s,  8/3s} {
        \draw (\x, 0.1) -- (\x, -0.1) node[above, yshift=2pt] {\xtext};
    }
    \draw[fill=black] (1,0.3) circle (0.2cm);
    \node[above] at (1,0.7) {$ $};
    \draw[fill=green] (5.5,0.3) circle (0.2cm);
    \node[above] at (5.5,0.5) {$ $};
    \node[align=center, above] at (6.5, 0) {\includegraphics[width=0.8cm]{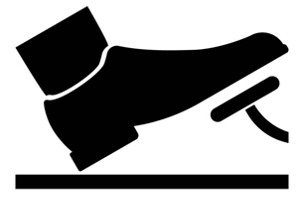}};
    \node[align=center, below] at (4, 0) {> x2 blocks, x30-x50 trials};
    \end{tikzpicture}
    & 64 & 7 h & 39 & 39 \\

    Singh2021 \cite{Singh2021} &  PD &
    \begin{tikzpicture}[baseline=(current bounding box.center), scale=0.8, every node/.style={scale=0.8}]
    \draw[|-|, thick] (0,0) -- (7.8,0);
    \foreach \x/\xtext in {2/1s, 7.8/8-20s} {
        \draw (\x, 0.1) -- (\x, -0.1) node[below] {\xtext};
    }
    \node[align=center, above] at (1, 0) {Instruction\\text:};
    \node[align=center, above] at (4.5, 0) {time-interval (3 or 7s)};
    \node[align=center, above] at (7, 0) {\includegraphics[width=1.2cm]{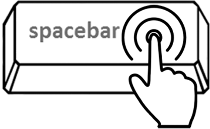}};
    \node[align=center, below] at (4, 0) {x80 trials, x4 blocks};
    \end{tikzpicture}
    & 64 & 58 h & 120 & 129 \\

TUAB \cite{TUHEEG} &  Abnormal & \multirow{4}{*}{
    \begin{tikzpicture}[baseline=(current bounding box.center)]
        \node (img) at (0,0) {\includegraphics[width=1.5cm]{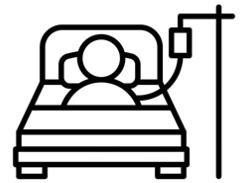}};
        \node[anchor=west, align=left] at (0.7,0) {\makecell[l]{\textit{Data collected from patients }\\\textit{in hospital beds during monitoring}}};
    \end{tikzpicture}
} & 19 & 47.5 d & 2,383 & 2,993 \\

TUEP \cite{tuep} & Epilepsy & & 19 & 26.3 d & 200 & 2,041 \\

TUAR \cite{tuar} & Artifact & & 19 & 4.2 d & 213 & 310 \\

CHB-MIT~ \cite{chbthesis} & Seizure &
    & 18 & 41 d & 23 & 686 \\
Sleep-Telemetry \cite{sleep_telemetry} &  \makecell[c]{Sleep\\Stages} &
    \begin{tikzpicture}[baseline=(current bounding box.center), scale=0.8, every node/.style={scale=0.8}]
    \draw[|-|, thick] (0,0) -- (3.9,0);
    \foreach \x/\xtext in {3.9/1night} {
        \draw (\x, 0.1) -- (\x, -0.1) node[below] {\xtext};
    }
    \node[align=center, above] at (2, 0) {\includegraphics[width=0.8cm]{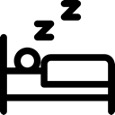}};
    \node[align=center, below] at (2, 0) {temazepam};
    \draw[|-|, thick] (4,0) -- (7.8,0);
    \foreach \x/\xtext in {7.8/1night} {
        \draw (\x, 0.1) -- (\x, -0.1) node[below] {\xtext};
    }
    \node[align=center, above] at (6, 0) {\includegraphics[width=0.8cm]{images/sleeping.jpg}};
    \node[align=center, below] at (6, 0) {placebo};
    \end{tikzpicture}
    & 3 & 15.8 d & 22 & 44 \\
    \bottomrule
\end{tabularx}
\smallskip
\caption{Clinical Datasets. PD stands for Parkinson's Disease, mTBI for mild Traumatic Brain Injury and OCD for Obsessive-Compulsive Disorder. \texticon{images/eyeopen.png} stands for eye open, \texticon{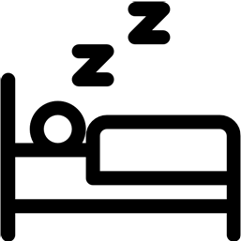} sleeping, \texticon{images/oddball.png} oddball stimulus, \texticon{images/simon_task.png} modified Simon task with the shapes and colors, \texticon{images/pedal_small.png} pedal press, \texticon{images/spacebar2.png} space bar press, \texticon{images/hospital2.png} hospital patient monitoring.}
    \label{tab:clinical_datasets_table} 
\end{table}
\noindent

\textbf{Datasets.} Our benchmark integrates 14 publicly available EEG datasets spanning multiple clinical paradigms, selected to reflect a wide range of real-world conditions. This includes heterogeneity in subject demographics, experimental protocols, clinical conditions, and recording setups and lengths. In this work, we focus on datasets (Table~\ref{tab:clinical_datasets_table}) that are primarily collected in medical contexts for diagnostic and monitoring purposes.

\textbf{Benchmarking Tasks}  To assess EEG foundation models in realistic healthcare settings, we define a suite of 11 clinical tasks (Table~\ref{tab:benchmarking_tasks}). All tasks are evaluated in a strict \textit{cross-subject} setting to ensure generalization beyond individual patients. The tasks cover both diagnostic classification and event detection. Diagnostic tasks involve full-length EEG recordings and include binary problems such as normal vs.\ abnormal or epileptic vs.\ non-epileptic, as well as disease-specific classifications for Parkinson’s disease (PD), obsessive–compulsive disorder (OCD), schizophrenia, and mild traumatic brain injury (mTBI).
As we have multiple datasets for PD, we additionally test the models on a dataset they were not trained on (held-out), for \textit{cross-dataset} evaluation.
Event-based tasks operate on shorter EEG segments and include seizure detection, artifact detection, and sleep stage classification. These tasks reflect the diversity of real-world clinical EEG use, with recordings ranging from minutes to hours, often involving uncontrolled cognitive states and substantial inter-subject variability. Together, they provide a challenging benchmark for evaluating whether foundation models can capture clinically meaningful neural representations.

\paragraph{Metrics} We report \textit{balanced accuracy} and \textit{weighted F1 score} as our primary metrics, computed on the fixed test splits. These metrics are chosen for their robustness to class imbalance and wide adoption in EEG decoding literature, respectively.

\begin{table}
\centering
\begin{minipage}[t]{.4\textwidth}
\begin{tabular}[t]{p{1.5cm} R{2.5cm} R{1.5cm}}
\toprule
\textbf{Task} & \textbf{Class} & \textbf{\#Samples} \\ 
\midrule
\multirow{2}{*}{Abnormal}              & Abnormal & 1,472                \\ 
& Normal & 1,521 \\
\hline
\multirow{2}{*}{Epilepsy}              & Epilepsy & 1,651                \\ 
& No Epilepsy & 390 \\
\hline
\multirow{2}{*}{PD (All)}           & Parkinson's & 266                  \\ 
& Control & 157 \\
\hline
\multirow{2}{*}{PD (Held-Out)}         & Parkinson's & 194                  \\ 
& Control & 115 \\
\hline
\multirow{2}{*}{OCD}                   & OCD & 22               \\ 
& Control & 24 \\
\hline
\multirow{2}{*}{mTBI}                  & mTBI & 104                  \\ 
& Control & 73 \\
\hline
\multirow{2}{*}{Schizophrenia}         & Schizophrenia & 45      \\ 
& Control & 31 \\[-.1cm]
\bottomrule
\end{tabular}
\end{minipage}
\hfill
\begin{minipage}[t]{.48\textwidth}
\begin{tabular}[t]{p{1.5cm} R{2.5cm} R{1.5cm}}
\toprule
\textbf{Task} & \textbf{Class} & \textbf{\#Samples} \\ 
\midrule
\multirow{2}{*}{Binary Artifact} & No Event & 216,466 \\ 
& Artifact & 141,102 \\
\hline
\multirow{6}{*}{\makecell{Multiclass\\Artifact}} & No Event & 216,466 \\ 
& Eye Movement & 34,480 \\
& Muscle Artifact & 60,763 \\
& Electrode Artifact & 41,652 \\
& Chewing & 3,964 \\
& Shivers & 243 \\
\hline
\multirow{6}{*}{\makecell{Sleep\\Stages}} & Unknown Phase & 82,734 \\ 
& Wake Phase & 132,746 \\
& N1 Phase & 109,590 \\
& N2 Phase & 595,530 \\
& N3 \& N4 Phase & 192,450 \\
& REM Phase & 250,470 \\
\hline
\multirow{2}{*}{Seizure} & No Seizure & 3,515,547 \\ 
& Seizure & 11,525 \\[-.1cm]
\bottomrule
\end{tabular}
\end{minipage}
\smallskip
\caption{Overview of the benchmarking tasks, showing the classes each task distinguishes.
\#Samples denotes the number of samples of each class.
Here, PD stands for Parkinson's Disease, OCD for Obsessive-Compulsive Disorder and mTBI for mild Traumatic Brain Injury.}\label{tab:benchmarking_tasks}
\end{table}

Our benchmarking framework supports both standard machine learning models that rely on handcrafted features and modern foundation models trained on large-scale EEG data.

\paragraph{Standard ML Models} As classical baselines, we use Linear Discriminant Analysis (LDA) and Support Vector Machines (SVM) applied to handcrafted features. All signals are resampled to 200 Hz, restricted to the subset of channels common across datasets, and truncated to a consistent length. Features are extracted with the Brainfeatures toolbox, which provides a comprehensive set of time- and frequency-domain measures. These include spectral power across standard frequency bands (delta–gamma), statistical moments, and complexity metrics such as entropy and fractal dimension. Transformations such as the continuous and discrete wavelet transforms (CWT, DWT) and the discrete Fourier transform (DFT) are applied to capture rich temporal–spectral dynamics. Brainfeatures also supports epoching and aggregation, enabling it to handle recordings of arbitrary length and generate consistent session-level representations, which is critical for heterogeneous clinical EEG.  

\paragraph{Foundation Models} We also evaluate foundation models that learn EEG representations directly from raw signals: BENDR~\cite{bendr}, Neuro-GPT~\cite{cui2024neurogpt}, and LaBraM~\cite{jiang2024large}. For all, signals are bandpass filtered (0.1–75 Hz), notch filtered to suppress line noise, and resampled to 200 Hz.  BENDR and Neuro-GPT require a fixed channel set; for each dataset, we hence retain the pretrained channels, zero-pad missing ones, and map datasets with incompatible layouts where necessary. We prepend every released model to a linear classification head and fine-tune the entire model.  For Neuro-GPT, following the authors’ findings, we fine-tune only the encoder.  LaBraM is more flexible with respect to channel configurations and can thus accommodate heterogeneous datasets. We use the publicly released LaBraM-Base and fine-tune it end-to-end.  Since clinical recordings often span minutes to hours, exceeding the input limits of these models (4–60 s), we segment each recording into non-overlapping chunks, encode them separately, and average embeddings before classification. This allows scalable inference on long-duration EEG while retaining global context. More details on how we used these models can be found in appendix \ref{sec:detailed_models}.

\section{Experiments}

We evaluate the generalization performance of both classical and foundation models across the EEG decoding tasks defined in our benchmark. Each experiment was repeated five times with different random seeds, requiring 270 hours on an A100 GPU and 16 AMD EPYC 7742 CPUs in total. We report the mean and standard deviation of the balanced accuracy scores in Table~\ref{tab:balanced_accuracies_results}. Corresponding weighted F1 scores are provided in Table~\ref{tab:f1_results} in the appendix for completeness.

\begin{table}[ht]

\centering
\begin{tabular}{l| C{1.2cm} | C{1.15cm} | C{2.1cm} | C{2.1cm} | C{2.1cm}}
\toprule
\textbf{Task} & \textbf{SVM} & \textbf{LDA} & \textbf{BENDR} & \textbf{Neuro-GPT} & \textbf{LaBraM} \\
\midrule
Abnormal & 0.722 & 0.677 & 0.717 ± .003 & 0.696 ± .005 & \textbf{0.838 ± .011} \\
Epilepsy & 0.531 & 0.531 & \textbf{0.740 ± .015} & 0.734 ± .010 & 0.565 ± .017 \\
PD (All) & 0.648 & 0.658 & 0.529 ± .009 & \textbf{0.687 ± .000} & 0.656 ± .025 \\
PD (Held-Out) & 0.596 & 0.654 & 0.615 ± .038 & \textbf{0.673 ± .000} & 0.673 ± .038 \\
OCD & 0.633 & 0.717 & 0.513 ± .051 & 0.703 ± .082 & \textbf{0.740 ± .044} \\
mTBI & 0.626 & \textbf{0.813} & 0.640 ± .093 & 0.646 ± .000 & 0.740 ± .173 \\
Schizophrenia & \textbf{0.679} & 0.547 & 0.471 ± .055 & 0.545 ± .042 & 0.543 ± .045 \\
Binary Artifact & 0.745 & 0.705 & 0.535 ± .003 & 0.711 ± .004 & \textbf{0.756 ± .007} \\
Multiclass Artifact & \textbf{0.437} & 0.325 & 0.192 ± .002 & 0.226 ± .006 & 0.430 ± .015 \\
Sleep Stages & 0.652 & \textbf{0.671} & 0.169 ± .001 & 0.166 ± .003 & 0.192 ± .001 \\
Seizure & 0.572 & 0.529 & 0.501 ± .001 & 0.500 ± .000 & \textbf{0.588 ± .011} \\
\bottomrule
\end{tabular}
\smallskip
\caption{Balanced Accuracy scores achieved by all models across evaluated tasks.
Features for SVM or LDA were extracted using the Brainfeatures toolbox.
For foundation models, we always report ``mean ± standard-deviation'' among the five runs.
}
\label{tab:balanced_accuracies_results}
\end{table}

We can see in Table~\ref{tab:balanced_accuracies_results} that LaBraM frequently achieved the highest or competitive performance across a range of tasks. For example, it reached a balanced accuracy of 0.838 in abnormal EEG detection — substantially outperforming all other models — and delivered strong results on
Obsessive-Compulsive Disorder (OCD) classification (0.740).
These results demonstrate LaBraM’s ability to model long, noisy, and heterogeneous EEG recordings, even under varied clinical conditions.
Interestingly, epilepsy detection presented an exception. Despite LaBraM’s strong performance on many other clinical tasks, it underperformed here with only 0.565 balanced accuracy, while BENDR and Neuro-GPT achieved much stronger results (0.740 and 0.734, respectively). Given the highly imbalanced nature of the epilepsy dataset, this suggests that LaBraM may be more sensitive to label imbalance or more prone to overfitting in such contexts. 
In the mild traumatic brain injury (mTBI) task, a simple classical method — LDA — surprisingly outperformed all foundation models, reaching a balanced accuracy of 0.813. This result highlights that in low-data regimes, classical models with strong inductive biases and low capacity can remain not only viable but also superior. All foundation models struggled on this task, likely due to their high capacity and lack of regularization in small-sample settings. In schizophrenia classification, SVM was the best-performing model (0.679), with all foundation models, including Neuro-GPT (0.545) and LaBraM (0.543), falling behind. None of the deep models learned meaningful patterns in this task 
— likely due to the subtle and distributed nature of schizophrenia-related EEG markers, which can be difficult to capture without tailored inductive structure or more specialized training data.
Throughout the multi-label tasks, we observe that SVM, LDA and LaBraM by and large yielded the highest scores, 
with at least one of the baseline models never being more than 2\% worse than LaBraM.
However, for sleep stages, the baselines yielded stable results around 0.66 balanced accuracy, while the foundation models severely struggled, with LaBraM -- the best among them -- achieving a score of merely 0.192. The inability of BENDR and Neuro-GPT to learn anything beyond random guessing for the sleep stages and seizure tasks might suggest that these models are not capable of using previously unseen channels in a meaningful way, as the datasets included in each task 
have no channels in common with either model. Despite 
not having been trained on the datasets' channels, LaBraM was able to advance to at least a few percentage points above random guesses, indicating a greater flexibility with respect to channels. 

\section{Conclusion}

As EEG foundation models continue to emerge, robust and clinically meaningful evaluation becomes increasingly important. In this work, we introduce a unified benchmarking framework for EEG decoding in healthcare, with a focus on diagnostic and event-based clinical tasks. By standardizing access to 14 publicly available datasets and defining 11 diverse classification tasks, our benchmark provides a transparent and extensible platform for evaluating models in realistic clinical settings.

\noindent Our contributions are threefold: \textit{(i) Standardized and Extensible Framework}: A unified platform that parses clinical EEG datasets into a common format, applies minimal preprocessing, and supports both classical and modern model evaluation. \textit{(ii) Well-defined Clinical Tasks:} A curated suite of 11 diagnostic and event-based tasks spanning a range of neurological and psychiatric conditions. \textit{(iii) Comprehensive Performance Evaluation:} Side-by-side comparisons of classical baselines and EEG foundation models to assess generalization across subjects and clinical conditions.

\paragraph{Limitations and Future Work.}
While our primary focus has been on building and validating the benchmarking framework, the current model suite includes only a few standard ML pipelines and foundation models. Many specialized models have been developed for individual tasks (e.g., seizure detection, sleep staging, schizophrenia diagnosis), and integrating these will provide a more complete view of model capabilities and limitations. We plan to continuously expand the benchmark to include more datasets, tasks, and models and encourage the community to contribute evaluations of emerging EEG models under this unified setup. By providing a rigorous, transparent, and extensible benchmark, we hope to foster more reliable progress in EEG decoding and support the development of foundation models that truly generalize across the diverse landscape of EEG data in clinical setups.

\clearpage

\bibliographystyle{unsrt}
\bibliography{references}

@article{medpalm2,
  title={Toward Expert-Level Medical Question Answering with Large Language Models},
  author={Singhal, Karan and Tu, Tao and Gottweis, Juraj and Sayres, Rory and Wulczyn, Ellery and Amin, Mohamed and Hou, Le and Clark, Kevin and Pfohl, Stephen R and Cole-Lewis, Heather and others},
  journal={Nature Medicine},
  volume={31},
  number={3},
  pages={943--950},
  year={2025}
}

@misc{healthbench,
      title={HealthBench: Evaluating Large Language Models Towards Improved Human Health}, 
      author={Rahul K. Arora and Jason Wei and Rebecca Soskin Hicks and Preston Bowman and Joaquin Quiñonero-Candela and Foivos Tsimpourlas and Michael Sharman and Meghan Shah and Andrea Vallone and Alex Beutel and Johannes Heidecke and Karan Singhal},
      year={2025},
      eprint={2505.08775},
      archivePrefix={arXiv},
      primaryClass={cs.CL}
}

@misc{wei2022emergent,
      title={Emergent Abilities of Large Language Models}, 
      author={Jason Wei and Yi Tay and Rishi Bommasani and Colin Raffel and Barret Zoph and Sebastian Borgeaud and Dani Yogatama and Maarten Bosma and Denny Zhou and Donald Metzler and Ed H. Chi and Tatsunori Hashimoto and Oriol Vinyals and Percy Liang and Jeff Dean and William Fedus},
      year={2022},
      eprint={2206.07682},
      archivePrefix={arXiv},
      primaryClass={cs.CL}
}

@misc{awais2023foundational,
      title={Foundational Models Defining a New Era in Vision: A Survey and Outlook}, 
      author={Muhammad Awais and Muzammal Naseer and Salman Khan and Rao Muhammad Anwer and Hisham Cholakkal and Mubarak Shah and Ming-Hsuan Yang and Fahad Shahbaz Khan},
      year={2023},
      eprint={2307.13721},
      archivePrefix={arXiv},
      primaryClass={cs.CV}
}

@misc{huang2023masked,
      title={Masked Autoencoders that Listen}, 
      author={Po-Yao Huang and Hu Xu and Juncheng Li and Alexei Baevski and Michael Auli and Wojciech Galuba and Florian Metze and Christoph Feichtenhofer},
      year={2023},
      eprint={2207.06405},
      archivePrefix={arXiv},
      primaryClass={cs.SD}
}

@ARTICLE{bendr,
  
AUTHOR={Kostas, Demetres and Aroca-Ouellette, Stéphane and Rudzicz, Frank},   
	 
TITLE={BENDR: Using Transformers and a Contrastive Self-Supervised Learning Task to Learn From Massive Amounts of EEG Data},      
	
JOURNAL={Frontiers in Human Neuroscience},      
	
VOLUME={15},           
	
YEAR={2021},      
	  
URL={https://www.frontiersin.org/articles/10.3389/fnhum.2021.653659},       
	
DOI={10.3389/fnhum.2021.653659},      
	
ISSN={1662-5161}
}

@misc{cui2024neurogpt,
      title={Neuro-GPT: Towards A Foundation Model for EEG}, 
      author={Wenhui Cui and Woojae Jeong and Philipp Thölke and Takfarinas Medani and Karim Jerbi and Anand A. Joshi and Richard M. Leahy},
      year={2024},
      eprint={2311.03764},
      archivePrefix={arXiv},
      primaryClass={cs.LG}
}

@misc{jiang2024large,
      title={Large Brain Model for Learning Generic Representations with Tremendous EEG Data in BCI}, 
      author={Wei-Bang Jiang and Li-Ming Zhao and Bao-Liang Lu},
      year={2024},
      eprint={2405.18765},
      archivePrefix={arXiv},
      primaryClass={cs.LG}
}

@article{CAVANAGH2018,
    title = {Diminished EEG habituation to novel events effectively classifies Parkinson’s patients},
    journal = {Clinical Neurophysiology},
    volume = {129},
    number = {2},
    pages = {409--418},
    year = {2018},
    issn = {1388-2457},
    doi = {https://doi.org/10.1016/j.clinph.2017.11.023},
    url = {https://www.sciencedirect.com/science/article/pii/S1388245717311719},
    author = {James F. Cavanagh and Praveen Kumar and Andrea A. Mueller and Sarah Pirio Richardson and Abdullah Mueen},
}

@article{ALBRECHT2019131,
title = {Increased conflict-induced slowing, but no differences in conflict-induced positive or negative prediction error learning in patients with schizophrenia},
journal = {Neuropsychologia},
volume = {123},
pages = {131--140},
year = {2019},
note = {Cognitive Effort},
issn = {0028-3932},
doi = {https://doi.org/10.1016/j.neuropsychologia.2018.04.031},
url = {https://www.sciencedirect.com/science/article/pii/S0028393218301726},
author = {Matthew A. Albrecht and James A. Waltz and James F. Cavanagh and Michael J. Frank and James M. Gold},
}

@article{Singh2018,
  author    = {Singh, A. and Richardson, S. P. and Narayanan, N. and Cavanagh, J. F.},
  title     = {Mid-frontal theta activity is diminished during cognitive control in Parkinson's disease},
  journal   = {Neuropsychologia},
  volume    = {117},
  pages     = {113--122},
  year      = {2018},
  month     = aug,
  doi       = {10.1016/j.neuropsychologia.2018.05.020},
  pmid      = {29802866},
  pmcid     = {PMC6524769},
  note      = {Epub 2018 May 23}
}

@article{Brown2020,
  author    = {Brown, D. R. and Richardson, S. P. and Cavanagh, J. F.},
  title     = {An EEG marker of reward processing is diminished in Parkinson's disease},
  journal   = {Brain Research},
  volume    = {1727},
  year      = {2020},
  month     = jan,
  doi       = {10.1016/j.brainres.2019.146541},
  pmid      = {31704082},
  note      = {Epub 2019 Nov 5}
}

@article{Gruendler2009,
  author    = {Gründler, T. O. and Cavanagh, J. F. and Figueroa, C. M. and Frank, M. J. and Allen, J. J.},
  title     = {Task-related dissociation in ERN amplitude as a function of obsessive-compulsive symptoms},
  journal   = {Neuropsychologia},
  volume    = {47},
  number    = {8-9},
  pages     = {1978--1987},
  year      = {2009},
  month     = jul,
  doi       = {10.1016/j.neuropsychologia.2009.03.010},
  pmid      = {19428431},
  pmcid     = {PMC2680784},
  note      = {Epub 2009 Mar 17}
}

@article{Cavanagh2019mtbi,
  author    = {Cavanagh, J. F. and Wilson, J. K. and Rieger, R. E. and Gill, D. and Broadway, J. M. and Story Remer, J. H. and Fratzke, V. and Mayer, A. R. and Quinn, D. K.},
  title     = {ERPs predict symptomatic distress and recovery in sub-acute mild traumatic brain injury},
  journal   = {Neuropsychologia},
  volume    = {132},
  pages     = {107--125},
  year      = {2019},
  month     = sep,
  doi       = {10.1016/j.neuropsychologia.2019.107125},
  pmid      = {31228481},
  pmcid     = {PMC6702033},
  note      = {Epub 2019 Jun 19}
}

@article{SINGH2020,
    title = {Frontal theta and beta oscillations during lower-limb movement in Parkinson’s disease},
    journal = {Clinical Neurophysiology},
    volume = {131},
    number = {3},
    pages = {694--702},
    year = {2020},
    issn = {1388-2457},
    doi = {https://doi.org/10.1016/j.clinph.2019.12.399},
    url = {https://www.sciencedirect.com/science/article/pii/S1388245720300092},
    author = {Arun Singh and Rachel C. Cole and Arturo I. Espinoza and Darin Brown and James F. Cavanagh and Nandakumar S. Narayanan},
}

@article{Singh2021,
  author    = {Singh, A. and Cole, R. C. and Espinoza, A. I. and Evans, A. and Cao, S. and Cavanagh, J. F. and Narayanan, N. S.},
  title     = {Timing variability and midfrontal $\sim$4 Hz rhythms correlate with cognition in Parkinson's disease},
  journal   = {NPJ Parkinson's Disease},
  volume    = {7},
  number    = {1},
  pages     = {14},
  year      = {2021},
  month     = feb,
  doi       = {10.1038/s41531-021-00158-x},
  pmid      = {33589640},
  pmcid     = {PMC7884691},
  note      = {Published 2021 Feb 15}
}

@ARTICLE{TUHEEG,
    AUTHOR={Obeid, Iyad  and Picone, Joseph },
    TITLE={The Temple University Hospital EEG Data Corpus},
    JOURNAL={Frontiers in Neuroscience},
    VOLUME={10},
    YEAR={2016},
    URL={https://www.frontiersin.org/journals/neuroscience/articles/10.3389/fnins.2016.00196},
    DOI={10.3389/fnins.2016.00196},
    ISSN={1662-453X},
}

@inproceedings{tuep,
    author = {Veloso, L. and McHugh, J. and von Weltin, Eva and Lopez, Sebas and Obeid, I. and Picone, Joseph},
    year = {2017},
    month = {12},
    pages = {1--3},
    title = {Big data resources for EEGs: Enabling deep learning research},
    doi = {10.1109/SPMB.2017.8257044}
}

@booklet{whoepilepsy,
    title = {Epilepsy: A public health imperative},
    year = {2019},
    author = {World Health Organization},
    note = {Geneva},
}

@mastersthesis{abnormal,
    author = {López de Diego, Silvia},
    title = {Automated Interpretation of Abnormal Adult Electroencephalograms},
    school = {Temple University},
    year = {2017}
}

@inproceedings{tuar,
  title={The Temple University Artifact Corpus: An Annotated Corpus of EEG Artifacts},
  author={Hamid, Ahmed and Gagliano, Katherine and Rahman, Safwanur and Tulin, Nikita and Tchiong, Vincent and Obeid, Iyad and Picone, Joseph},
  booktitle={2020 IEEE Signal Processing in Medicine and Biology Symposium (SPMB)},
  pages={1--4},
  year={2020},
  organization={IEEE}
}

@phdthesis{chbthesis,
  title={Application of Machine Learning to Epileptic Seizure Onset Detection and Treatment},
  author={Shoeb, Ali Hossam},
  year={2009},
  school={Massachusetts Institute of Technology},
  url = {https://doi.org/10.13026/C2K01R},
}

@article{sleep_telemetry,
    author = {Mourtazaev, M. S. and Kemp, B. and Zwinderman, A. H. and Kamphuisen, H. A.},
    title = {Age and gender affect different characteristics of slow waves in the sleep EEG},
    journal = {Sleep},
    volume={18},
    pages = {557--564},
    year = {1995}
}

\newpage
\appendix

\section{EEG-Bench Interface}\label{sec:interface}
We have made the code of our benchmarking tool publicly available at \url{https://github.com/ETH-DISCO/EEG-Bench}, licensed under the GNU GPL v3.0 license or later.
\subsection*{Running the benchmark}
To run the benchmark, create the conda environment \texttt{eeg\_bench} via \begin{verbatim}
    conda env create -f environment.yml
    conda activate eeg_bench
\end{verbatim}
Then, configure the paths to your local storage in \texttt{eeg\_bench/config.json} and run the benchmark via \begin{verbatim}
    python benchmark_console.py --all
\end{verbatim}

\subsection*{Dataset Accessibility and Distribution}
\label{sec:distribution}

A core design principle of EEG-Bench is ease of use: To foster widespread adoption and reproducibility, we aim to abstract away the often cumbersome process of data acquisition. Thus, we implemented automated download mechanisms for public datasets using custom scripts integrated into our codebase. This ensures that users can seamlessly access all of the 14 datasets with minimal effort and (most of the time) no need for manual intervention.

Notably, datasets hosted by the Temple University Hospital (TUAB, TUEP, TUAR) require acceptance of a data use agreement (DUA) via the NEDC portal; once accepted, they can be automatically retrieved using our tools.

The table below summarizes dataset accessibility across all 14 benchmark datasets.

\begin{table}[h]
\centering
\begin{tabular}{@{}lcc@{}}
\toprule
\textbf{Dataset Name} & \textbf{Access Type} & \textbf{Notes / Requirements} \\
\midrule
TUAB               & Automatic  & DUA via NEDC; automatic download afterward \\
TUEP               & Automatic  & DUA via NEDC; automatic download afterward \\
TUAR               & Automatic  & DUA via NEDC; automatic download afterward \\
CHB-MIT            & Automatic  &  \\
Sleep-Telemetry    & Automatic  &  \\
Cavanagh2018a      & Automatic  &  \\
Cavanagh2018b      & Automatic  &  \\
Cavanagh2019       & Automatic  &  \\
Singh2018          & Automatic  &  \\
Singh2020          & Automatic  &  \\
Singh2021          & Automatic  &  \\
Brown2020          & Automatic  &  \\
Gruendler2009      & Automatic  &  \\
Albrecht2019       & Automatic  &  \\
\bottomrule
\end{tabular}
\smallskip
\caption{Dataset Accessibility Overview}
\label{tab:dataset_access}
\end{table}

\subsection*{Adding Your Own Dataset}
Our benchmarking code allows users to easily add their own datasets to the benchmark. Base classes are available for both, clinical and BCI datasets.

To add your dataset, follow these steps:

\begin{enumerate}
    \item Place your dataset class in \texttt{eeg\_bench/datasets/bci/} or \texttt{eeg\_bench/datasets/clinical/}.
    
    \item Inherit from \texttt{BaseBCIDataset} or \texttt{BaseClinicalDataset}.
    
    \item Implement the following methods:
    \begin{enumerate}
        \item \texttt{\_download}: Either download the dataset automatically or provide instructions for the user to do so manually. Pay attention that, if possible, \texttt{\_download} does not re-download the dataset if it already exists locally.
        
        \item \texttt{load\_data}: This method should populate the following attributes:
        \begin{itemize}
            \item \texttt{self.data}: \texttt{np.ndarray} or \texttt{List[BaseRaw]} with shape \((n_{\text{samples}}, n_{\text{channels}}, n_{\text{sample length}})\)
            \item \texttt{self.labels}: \texttt{np.ndarray} or \texttt{List[str]} with shape \((n_{\text{samples}},)\), or $(n_\text{samples}, n_\text{multi\_labels})$ for multi-label datasets
            \item \texttt{self.meta}: A dictionary that must contain at least \texttt{name}, \texttt{sampling\_frequency} and \texttt{channel\_names}
        \end{itemize}
        
        \item If your dataset contains classes not yet defined in the enums \texttt{enums.BCIClasses} or \texttt{enums.ClinicalClasses}, please add them accordingly.
        \item For multi-label datasets, you currently also have to add your dataset name to the
        
        \hs{.4}\texttt{elif dataset\_name in [<MULTILABEL\_DATASET\_NAMES>]:}
        
        clause in \path{eeg_bench/models/clinical/brainfeatures/feature_extraction_2.py:_prepare_data_cached()}.

        \item To speed up further runs of the \texttt{load\_data} function, implement caching as in the existing dataset classes.
        
        \item All EEG signals should be standardized to the microvolt (\(\mu\text{V}\)) scale. To reduce memory usage and computational overhead, signals with a sampling rate greater than 250~Hz are typically resampled to 250~Hz.
    \end{enumerate}
\end{enumerate}

\subsection*{Adding Your Own Task}

Tasks are the central organizing principle of the benchmark, encapsulating paradigms, datasets, prediction classes, subject splits (i.e., training and test sets), and evaluation metrics. Each task class implements a \texttt{get\_data()} method that returns training or testing data, along with the corresponding labels and metadata. These predefined splits ensure evaluation consistency and facilitate reproducibility. The tasks are divided into \textbf{Clinical} and \textbf{BCI} categories.

Each task defines:
\begin{itemize}
    \item The datasets to be used
    \item Training and testing subject splits
    \item Target classes
    \item Evaluation metrics
\end{itemize}

To add your own task:
\begin{itemize}
    \item For BCI tasks, add your class to \texttt{tasks/bci/} and inherit from \texttt{AbstractBCITask}
    \item For Clinical tasks, add your class to \texttt{tasks/clinical/} and inherit from \texttt{AbstractClinicalTask}
\end{itemize}

You must implement the \texttt{get\_data()} method to return training or testing splits along with data, labels, and metadata.

For multi-label tasks, you must also add its name to the \texttt{get\_multilabel\_tasks()} method in \texttt{eeg\_bench/utils/utils.py}. Additionally, if you have special channel requirements, you might also want to add an
\begin{verbatim}
elif task_name == <YOUR_TASK_NAME>:
    t_channels = <YOUR_CHANNEL_LIST>
\end{verbatim}
clause to \texttt{\_prepare\_data\_cached()} in \path{eeg_bench/models/clinical/brainfeatures/feature_extraction_2.py}.


\subsection*{Add Your Own Model}

To integrate a new model into the benchmark, implement the \texttt{AbstractModel} interface and place your implementation in the appropriate directory:
\begin{itemize}
    \item \texttt{models/bci/} for Motor Imagery (BCI) models
    \item \texttt{models/clinical/} for Clinical models
\end{itemize}

\subsubsection*{Required Methods}

Your model must implement the following methods:

\begin{verbatim}
def fit(self, X: List[np.ndarray | List[BaseRaw]],
              y: List[np.ndarray | List[str]],
              meta: List[Dict]) -> None:
    # Each list entry corresponds to one dataset
    pass

def predict(self, X: List[np.ndarray | List[BaseRaw]],
                  meta: List[Dict]) -> np.ndarray:
    # Predict on each dataset separately, return concatenated predictions
    pass
\end{verbatim}

\subsection*{Running Your Model}

To run your model, register it in \texttt{benchmark\_console.py} and execute the following command:

\begin{verbatim}
python benchmark_console.py --model mymodel --task <YOUR_DESIRED_TASK>
\end{verbatim}

\section{Models}\label{sec:detailed_models}

In this section, we provide a more detailed description of all the baselines and the foundation models used for this paper.

\subsection{Standard ML Baselines}

To establish strong baselines for the benchmarking tasks, we employed classical machine learning pipelines that are well-established in EEG research. Specifically, we used Linear Discriminant Analysis (LDA) and Support Vector Machines (SVM), which operate on handcrafted features derived from the EEG signal. These models are known for their robustness in low-data regimes and their interpretability, making them a natural starting point for benchmarking. 
To ensure uniformity across datasets, for standard ML baselines, all EEG signals are resampled to 200 Hz. Since datasets differ in both channel layout and recording duration, for standard ML models, we restrict our analysis to the subset of channels common across datasets and truncate signals to a consistent length. Following this, we apply the feature extraction techniques described below.


\paragraph{Feature Extraction in Clinical Tasks:}  
For our clinical classification tasks, which include long-duration recordings and full-recording-level predictions, we utilize the Brainfeatures toolbox\footnote{Code available at \url{https://github.com/TNTLFreiburg/brainfeatures}}. This open-source tool provides a comprehensive set of EEG features derived from both time and frequency domains. It applies a variety of transformations—including the continuous wavelet transform (CWT), discrete wavelet transform (DWT), and discrete Fourier transform (DFT)—to extract rich representations of EEG dynamics. Extracted features include spectral power across standard frequency bands (delta: 0.5–4 Hz, theta: 4–8 Hz, alpha: 8–13 Hz, beta: 13–30 Hz, and gamma: 30–100 Hz), statistical moments (mean, variance, higher-order moments), as well as complexity metrics such as entropy and fractal dimension. Crucially, Brainfeatures includes robust mechanisms for epoching and aggregating features over time, enabling it to process EEG recordings of arbitrary length and generate consistent, session-level feature representations. This makes it particularly suitable for clinical applications where recording conditions are more heterogeneous.

\subsection{Large Pretrained Models}

In addition to classical pipelines, our benchmark incorporates large pretrained deep learning models designed to learn generalizable EEG representations directly from raw signals with minimal preprocessing. We evaluate three recent architectures: BENDR, Neuro-GPT, and LaBraM.

\paragraph{BENDR}
BENDR \cite{bendr} requires a fixed set of input channels. Therefore, we select only the subset of channels used during pretraining. If any of these channels are missing in a given dataset, we zero-pad the corresponding input. An exception to this rule comes into effect when \textit{all} channels are missing, as is the case in the Sleep-Telemetry and CHB-MIT datasets. In this case, we assign the dataset channels to arbitrary BENDR input channels. We apply a bandpass filter from 0.1 to 75.0 Hz and a notch filter to suppress power line artifacts. All signals are resampled to 200 Hz to match the model's expected input rate. We use the publicly released pretrained BENDR encoder and append a linear classification head. The entire model, including the encoder, is fine-tuned on each task.

\paragraph{Neuro-GPT}
Similar to BENDR, Neuro-GPT \cite{cui2024neurogpt} requires a fixed channel layout, so we apply the same preprocessing steps, including channel selection, bandpass and notch filtering, and resampling to 200 Hz. We utilize the pretrained model released by the authors and follow their recommended approach of fine-tuning the encoder only, as it was shown to yield the best performance in their experiments.

\paragraph{LaBraM}
LaBraM \cite{jiang2024large} is more flexible with respect to channel configurations, making it more suitable for heterogeneous EEG datasets. As with the other models, we apply a 0.1–75.0 Hz bandpass filter, a notch filter, and resample all signals to 200 Hz. We use the LaBraM-Base variant, which is the only version publicly released, and fine-tune it end-to-end on our tasks.

\paragraph{Handling Long Clinical Recordings}
Clinical datasets often contain continuous recordings spanning several minutes or hours, exceeding the input length limits of these models (ranging from 4 to 60 seconds, depending on the architecture). To address this, we divide each sample exceeding these length limits into \(N\) non-overlapping chunks of fixed duration. Each chunk is passed through the encoder, and the resulting embeddings are averaged across all chunks before being fed into the final classification layer. This strategy allows efficient and scalable inference over long-duration EEG data while preserving global context through embedding aggregation.

\newpage

\section{Weighted F1-Score Results and Task Details}
In the following table, we report the results of our experiments from Section 3 also using weighted F1 score.

\begin{table}[ht]
\centering
\begin{tabular}{l| C{1.2cm} | C{1.15cm} | C{2.1cm} | C{2.1cm} | C{2.1cm}}
\toprule
\textbf{Task} & \textbf{SVM} & \textbf{LDA} & \textbf{BENDR} & \textbf{Neuro-GPT} & \textbf{LaBraM} \\
\midrule
Abnormal & 0.720 & 0.680 & 0.722 ± .003 & 0.699 ± .005 & \textbf{0.842 ± .012} \\
Epilepsy & 0.613 & 0.593 & \textbf{0.709 ± .029} & 0.697 ± .016 & 0.647 ± .018 \\
PD (All) & 0.670 & 0.682 & 0.560 ± .016 & \textbf{0.724 ± .000} & 0.692 ± .021 \\
PD (Held-Out) & 0.662 & 0.707 & 0.683 ± .043 & 0.729 ± .000 & \textbf{0.742 ± .037} \\
OCD & 0.636 & 0.723 & 0.372 ± .127 & 0.681 ± .089 & \textbf{0.743 ± .042} \\
mTBI & 0.580 & \textbf{0.793} & 0.704 ± .093 & 0.683 ± .000 & 0.776 ± .152 \\
Schizophrenia & \textbf{0.681} & 0.533 & 0.421 ± .084 & 0.544 ± .041 & 0.463 ± .128 \\
Binary Artifact & \textbf{0.761} & 0.728 & 0.554 ± .003 & 0.723 ± .006 & 0.752 ± .009 \\
Multiclass Artifact & \textbf{0.683} & 0.643 & 0.430 ± .002 & 0.427 ± .012 & 0.624 ± .017 \\
Sleep Stages & 0.700 & \textbf{0.732} & 0.245 ± .001 & 0.231 ± .010 & 0.264 ± .002 \\
Seizure & 0.974 & 0.987 & 0.994 ± .000 & \textbf{0.995 ± .000} & 0.986 ± .003 \\
\bottomrule
\end{tabular}
\smallskip
\caption{Weighted F1-scores achieved by all models across evaluated tasks. Features for SVM or LDA were extracted using the Brainfeatures toolbox.}
\label{tab:f1_results}
\end{table}

Additionally, the following table specifies the datasets included in the evaluation set of each task.

\newcommand{\rot}[1]{\rotatebox{90}{#1}}

\begin{table}[htbp]
  \centering
  \begin{minipage}[b]{0.99\textwidth}
    \centering
    \begin{adjustbox}{max width=\linewidth}
    \begin{tabular}{l|ccccccccccc}
    \toprule
    \textbf{Dataset} & \rot{\textbf{Abnormal}} & \rot{\textbf{Epilepsy}} & \rot{\textbf{PD (All)}} & \rot{\textbf{PD (Held-Out)}} & \rot{\textbf{OCD}} & \rot{\textbf{mTBI}} & \rot{\textbf{Schizophrenia}} & \rot{\textbf{Binary Artifact}} & \rot{\textbf{Multiclass Artifact}} & \rot{\textbf{Sleep Stages}} & \rot{\textbf{Seizure}} \\
    \midrule
    TUAB              & \checkmark &   &   &    &    &    &    \\
    TUEP              &   & \checkmark &   &    &    &    &    \\
    Cavanagh2018a     &   &   & \checkmark &    &    &    &    \\
    Cavanagh2018b     &   &   & \checkmark &    &    &    &    \\
    Singh2018         &   &   & \checkmark &    &    &    &    \\
    Brown2020         &   &   & \checkmark &    &    &    &    \\
    Singh2020         &   &   & \checkmark & \checkmark  &  &  &  \\
    Singh2021         &   &   & \checkmark &    &    &    &    \\
    Gruendler2009     &   &   &   &   & \checkmark   &    &    \\
    Cavanagh2019      &   &   &   &   &    &  \checkmark  &    \\
    Albrecht2019      &   &   &   &   &    &    &  \checkmark  \\
    TUAR              &   &   &   &   &    &    &    & \checkmark & \checkmark \\
    Sleep-Telemetry   &   &   &   &   &    &    &    &    &    & \checkmark \\
    CHB-MIT           &   &   &   &   &    &    &    &    &    &    & \checkmark \\
    \bottomrule
    \end{tabular}
    \end{adjustbox}
    \smallskip
    \caption{Overview of clinical dataset inclusion. PD=Parkinson's disease, OCD=Obsessive-Compulsive Disorder, mTBI=mild Traumatic Brain Injury.}
  \end{minipage}
\end{table}

Lastly, Table~\ref{tab:detailed_tasks} gives more detailed information about each of the tasks that we defined.

\begin{table}
\centering
\begin{tabularx}{\textwidth}{p{3.3cm} R{2.5cm} R{1.5cm} R{1.5cm} R{2cm}}
\toprule
\textbf{Task} & \textbf{Class} & \multicolumn{1}{r}{\textbf{Train Samples}} & \multicolumn{1}{r}{\textbf{Test Samples}} &  \textbf{Sample Length} \\
\midrule
\multirow{2}{*}{Abnormal}              & Abnormal & 1,346                & 126                & \multirow{2}{*}{22.9 m} \\
& Normal & 1,371 & 150 \\
\midrule
\multirow{2}{*}{Epilepsy}              & Epilepsy & 1,384                 & 267                & \multirow{2}{*}{18.4 m} \\
& No Epilepsy & 268 & 122 \\
\midrule
\multirow{4}{*}{
\hs{-.35}\begin{tabularx}{\textwidth}{l l}
    \multirow{3}{*}{\makecell[l]{Parkinson's\\Disease\\(PD)}} & (All)\\\\
    & (Held-Out)\\
\end{tabularx}
}           & Parkinson's & 168                  & 98                  & \multirow{2}{*}{24.8 m} \\
& Control & 102 & 55 \\
& Parkinson's & 168                  & 26                  & \multirow{2}{*}{24.1 m} \\
& Control & 102 & 13 \\
\midrule
\multirow{2}{*}{\makecell[l]{Obsessive-Compulsive\\Disorder (OCD)}}                   & OCD & 17 & 5               & \multirow{2}{*}{22.6 m} \\
& Control & 18 & 6 \\
\midrule
\multirow{2}{*}{\makecell[l]{mild Traumatic\\Brain Injury (mTBI)}}                  & mTBI & 85                   & 19                  & \multirow{2}{*}{16.5 m} \\
& Control & 64 & 9 \\
\midrule
\multirow{2}{*}{Schizophrenia}         & Schizophrenia & 38 & 7      & \multirow{2}{*}{27.0 m} \\
& Control & 25 & 6 \\
\midrule

\multirow{2}{*}{Binary Artifact} & No Event & 160,879 & 55,587 & \multirow{2}{*}{16 s} \\
& Artifact & 107,457 & 33,645 \\
\midrule
\multirow{6}{*}{Multiclass Artifact} & No Event & 160,879 & 55,587 & \multirow{6}{*}{16 s} \\
& Eye Movement & 27,196 & 7,284 \\
& Muscle Artifact & 44,451 & 16,312 \\
& Electrode Artifact & 32,104 & 9,548 \\
& Chewing & 3,494 & 470 \\
& Shivers & 212 & 31 \\
\midrule
\multirow{6}{*}{Sleep Stages} & Unknown Phase & 65,794 & 16,940 & \multirow{6}{*}{16 s} \\
& Wake Phase & 101,546 & 31,200 \\
& N1 Phase & 84,300 & 25,290 \\
& N2 Phase & 443,070 & 152,460 \\
& N3 \& N4 Phase & 150,900 & 41,550 \\
& REM Phase & 191,190 & 59,280 \\
\midrule
\multirow{2}{*}{Seizure} & No Seizure & 2,548,702 & 966,845 & \multirow{2}{*}{16 s} \\
& Seizure & 8,514 & 3,011 \\
\bottomrule
\end{tabularx}
\smallskip
\caption{Detailed overview of the benchmarking tasks, displaying the number of samples per class and train/test subset, as well as the average sample length.}\label{tab:detailed_tasks}
\end{table}

\section{Datasets}

Clinical datasets reflect neural conditions that arise naturally or as a result of neurological or psychiatric disorders. These datasets are valuable for automating the detection of pathological events and conditions such as seizures, epilepsy, Parkinson’s disease, and schizophrenia, as well as non-pathological states like sleep stages.
They are often collected in hospital settings and, as a result, they tend to exhibit higher levels of noise and variability than, for example, BCI datasets recorded in a laboratory environment. While this introduces challenges for the model to learn in spite of this noise, it can also improve the robustness and generalizability of models trained on such data for real-world applications.
Below, we provide an overview of the clinical datasets included in our benchmark.

\subsection{Cavanagh2018a}
The Cavanagh2018a dataset originates from an EEG-based clinical study investigating neural habituation to novel auditory stimuli in Parkinson’s disease (PD) patients and matched controls \cite{CAVANAGH2018}. The primary goal was to evaluate whether EEG responses to novelty could serve as a biomarker for cognitive dysfunction in PD.

A total of 53 participants were involved: 25 individuals diagnosed with Parkinson’s disease (mean age: 69.68 ± 8.73 years; 16 male, 9 female) and 28 healthy age- and sex-matched control subjects. Each PD participant completed two sessions: one ON dopaminergic medication and one OFF (after 15 hours of medication withdrawal). Control participants completed a single session. All sessions were conducted at 9 AM to reduce circadian variability.

Participants underwent a three-stimulus auditory oddball task while EEG was recorded from 64 Ag/AgCl scalp electrodes placed according to the international 10/20 system. The task consisted of two blocks of 100 trials each, with three types of stimuli: frequent standard tones (440 Hz, 70\% of trials), infrequent target tones (660 Hz, 15\%), and novel sounds (15\%) taken from naturalistic audio recordings. Each stimulus lasted 200 ms, and the average task duration per session was approximately 12 minutes.

The participants were instructed to silently count the target tones and ignore standards and novels. This passive paradigm was designed to isolate cognitive processing without motor response confounds.

\begin{table}[!htbp]
\centering
\begin{tabular}{|c|c|}
\hline
Number of subjects & 53 (25 PD, 28 Control) \\
\hline
Sessions per subject (average) & 1.54 \\
\hline
Average session length & 16.4 min \\
\hline
EEG channels & 64 \\
\hline
Sampling rate & 500 Hz \\
\hline
Total recordings & 77 \\
\hline
Hardware & Brain Vision system \\
\hline
\end{tabular}
\smallskip
\caption{Summary of the Cavanagh2018a dataset}
\end{table}

\subsection{Cavanagh2018b}
The Cavanagh2018b dataset contains resting-state EEG recordings from the same group of people used in the Cavanagh2018a study \cite{CAVANAGH2018}. Participants included 28 individuals diagnosed with Parkinson’s disease and 28 age- and sex-matched control participants. Each subject underwent a 2-minute resting-state EEG recording session with eyes open. These recordings were collected in a seated posture, prior to or following the auditory oddball task, under the same EEG setup (64-channel cap, 500 Hz sampling rate).

This dataset serves as a complementary baseline condition for evaluating spontaneous brain dynamics in Parkinson’s disease. Though not central to the novelty task paradigm, resting-state data may be useful for future investigations into low-frequency oscillations or non-task-based classification approaches of Parkinson's disease.

\begin{table}[!htbp]
\centering
\begin{tabular}{|c|c|}
\hline
Number of subjects & 56 (28 PD, 28 Control) \\
\hline
Sessions per subject & 1 \\
\hline
Average session length & 2 min \\
\hline
EEG channels & 64 \\
\hline
Sampling rate & 500 Hz \\
\hline
Total recordings & 56 \\
\hline
Hardware & Brain Vision system \\
\hline
\end{tabular}
\smallskip
\caption{Summary of the Cavanagh2018b dataset}
\end{table}

\subsection{Albrecht2019}
The Albrecht2019 dataset originates from a study investigating reinforcement learning under cognitive conflict in individuals with schizophrenia (PSZ) and healthy controls \cite{ALBRECHT2019131}. The dataset includes both behavioral and EEG recordings collected during a modified Simon task, which introduces response conflict as an implicit cognitive cost during reinforcement learning.

A total of 78 participants took part in the study: 46 individuals with a DSM-IV diagnosis of schizophrenia or schizoaffective disorder, and 32 healthy controls. EEG data were recorded using a 64-channel Brain Vision system at a 1000 Hz sampling rate. Data were preprocessed and artifact-corrected using the EEGLAB pipeline, and ICA was applied for eye-blink removal. EEG epochs were extracted around stimulus and feedback events to capture conflict-evoked and prediction-error-related activity, particularly in the theta band (4–7 Hz).

Participants completed a reinforcement learning version of the Simon task. On each trial, a stimulus was associated with probabilistic reward or punishment outcomes, modulated by whether the trial involved a congruent or conflict-inducing response. This design enabled the dissociation of positive and negative prediction error (PE) learning biases under cognitive effort. A subsequent transfer phase assessed stimulus preferences to infer learning outcomes.

\begin{table}[!htbp]
\centering
\begin{tabular}{|c|c|}
\hline
Number of subjects & 78 (46 PSZ, 32 Controls) \\
\hline
Sessions per subject & 1 \\
\hline
Average session length & 40.3 min \\
\hline
EEG channels & 64 \\
\hline
Sampling rate & 1000 Hz \\
\hline
Total recordings & 76 \\
\hline
Hardware & Brain Vision system \\
\hline
\end{tabular}
\smallskip
\caption{Summary of the Albrecht2019 dataset}
\end{table}

\subsection{Singh2018}
The Singh2018 dataset includes EEG recordings collected during a cognitive control task from 28 individuals with Parkinson’s disease (PD) and 28 demographically matched healthy controls \cite{Singh2018}. Each participant completed a modified Simon reaction-time task designed to elicit response conflict and error-related cognitive control processes. PD patients participated in two sessions (ON and OFF dopaminergic medication), spaced one week apart, while controls participated in a single session.

EEG was recorded from 64 scalp electrodes using a Brain Vision system at a sampling rate of 500 Hz.

\begin{table}[!htbp]
\centering
\begin{tabular}{|c|c|}
\hline
Number of subjects & 56 (28 PD, 28 Controls) \\
\hline
Sessions per subject & 1 \\
\hline
Average session length & 118 min \\
\hline
EEG channels & 64 \\
\hline
Sampling rate & 500 Hz \\
\hline
Total recordings & 55 \\
\hline
Hardware & Brain Vision system \\
\hline
\end{tabular}
\smallskip
\caption{Summary of the Singh2018 dataset}
\end{table}

\subsection{Brown2020}
The Brown2020 dataset comprises EEG recordings from a reinforcement learning task aimed at assessing reward processing in individuals with Parkinson’s disease (PD) and healthy controls \cite{Brown2020}. A total of 56 participants took part: 28 individuals diagnosed with PD and 28 age- and sex-matched control participants. Each PD participant completed two sessions (ON and OFF dopaminergic medication), spaced one week apart. Control participants completed a single session.

Participants performed a reinforcement learning task involving probabilistic feedback. On each trial, a pair of colored stimuli was presented, with each stimulus associated with a predefined probability of reward. Conditions were manipulated along two dimensions: difficulty (90/10\% vs. 70/30\% reward probability) and volition (free choice vs. instructed choice). The EEG was time-locked to the feedback screen, allowing for the measurement of reward-related event-related potentials (ERPs).

EEG data were recorded using a 64-channel Brain Vision system at a sampling rate of 500 Hz.

\begin{table}[!htbp]
\centering
\begin{tabular}{|c|c|}
\hline
Number of subjects & 56 (28 PD, 28 Controls) \\
\hline
Sessions per subject & 1.5 \\
\hline
Average session length & 22.1 min \\
\hline
EEG channels & 64 \\
\hline
Sampling rate & 500 Hz \\
\hline
Total recordings & 84 \\
\hline
Hardware & Brain Vision system \\
\hline
\end{tabular}
\smallskip
\caption{Summary of the Brown2020 dataset}
\end{table}

\subsection{Gruendler2009}
The dataset Gruendler2009 \cite{Gruendler2009} originates from an EEG experiment that examined the relationship between obsessive–compulsive (OC) symptomatology and error-related brain activity. Participants were 46 undergraduate students selected based on their scores on the Obsessive–Compulsive Inventory-Revised (OCI-R), with groups categorized as high or low OC.

The experiment consisted of a flanker task to elicit ERNs from motor errors in a response conflict paradigm. Specifically, participants were shown a 5-letter string like ``QQQQQ'' or ``QQOQQ''. Within a 1-second time-window, they then had to press a left or right button, depending on the letter in the middle of the string.

The EEG data were recorded using a 64-channel EEG + 2-channel EOG + 1-channel EKG setup. Participants were excluded for poor EEG quality, failure to meet learning criteria, or inconsistent OCI-R group classification. Demographic and psychometric data (including the Beck Depression Inventory) were collected to control for confounds.

\begin{table}[!htbp]
\centering
\begin{tabular}{|c|c|}
\hline
Number of subjects & 46 \\
\hline
Sessions per subject & 1 \\
\hline
Average session length & 28.7 min \\
\hline
EEG channels & 64 \\
\hline
Sampling rate & 500 Hz \\
\hline
Total recordings & 46 \\
\hline
Hardware & NeuroScan SynAmps2 \\
\hline
\end{tabular}
\smallskip
\caption{Summary of the Gruendler2009 dataset}
\end{table}

\subsection{Cavanagh2019}
The Cavanagh2019 dataset includes EEG recordings collected during a 3-stimulus auditory oddball paradigm in participants with mild traumatic brain injury (mTBI) and matched healthy controls \cite{Cavanagh2019mtbi}. A total of 85 participants took part: 38 sub-acute mTBI patients (tested within 2 weeks post-injury), 24 healthy controls, and 23 chronic TBI patients (mild to moderate severity). Sub-acute mTBI and control participants completed two or three EEG sessions -- at 3–14 days and again after approximately 2 months -- while chronic TBI participants completed a single session.

The task involved 260 trials: 70\% standard tones (440 Hz), 15\% target tones (660 Hz), and 15\% novel naturalistic sounds. Stimuli were presented binaurally, and participants were instructed to count target tones while ignoring the others. EEG was recorded from 64 channels at a 500 Hz sampling rate.

\begin{table}[!htbp]
\centering
\begin{tabular}{|c|c|}
\hline
Number of subjects & 85 (38 sub-acute mTBI, 24 control, 23 chronic TBI) \\
\hline
Sessions per subject & 2 (mTBI, controls), 1 (chronic TBI) \\
\hline
Average session length & 22.1 min \\
\hline
EEG channels & 64 \\
\hline
Sampling rate & 500 Hz \\
\hline
Total recordings & 84 \\
\hline
Hardware & Brain Vision system \\
\hline
\end{tabular}
\smallskip
\caption{Summary of the Cavanagh2019 dataset}
\end{table}

\subsection{Singh2020}
The Singh2020 dataset contains EEG recordings collected during a lower-limb pedaling task designed to assess motor control in individuals with Parkinson’s disease (PD), with a particular focus on freezing of gait (FOG) symptoms \cite{SINGH2020}. A total of 39 participants were included: 13 PD patients with FOG (PDFOG+), 13 PD patients without FOG (PDFOG-), and 13 demographically matched healthy controls.

Participants completed a lower-limb motor task in which they pedaled a stationary cycle in response to a visual “GO” cue, designed to minimize fall risk and reduce EEG artifacts from movement. Each subject completed at least two blocks of either 30 or 50 trials, with PDFOG+ participants performing fewer trials due to symptom severity. A tri-axial accelerometer mounted on the ankle measured pedaling kinematics, including mean speed and time to peak acceleration.

EEG was recorded using a 64-channel cap with a sampling rate of 500 Hz.

\begin{table}[!htbp]
\centering
\begin{tabular}{|c|c|}
\hline
Number of subjects & 39 (13 PDFOG+, 13 PDFOG-, 13 Controls) \\
\hline
Sessions per subject & 1 \\
\hline
Average session length & 10.8 min \\
\hline
EEG channels & 64 \\
\hline
Sampling rate & 500 Hz \\
\hline
Total recordings & 39 \\
\hline
Hardware & Easycap Brain Products \\
\hline
\end{tabular}
\smallskip
\caption{Summary of the Singh2020 dataset}
\end{table}

\subsection{Singh2021}
The Singh2021 dataset contains EEG recordings collected during an interval timing task designed to study cognitive control in individuals with Parkinson’s disease (PD) \cite{Singh2021}. A total of 130 participants were recruited: 89 PD patients and 41 demographically matched healthy controls. Of these, usable EEG data were available for 83 PD patients and 37 controls, after excluding sessions with insufficient data or poor signal quality. Most PD patients (n = 80) completed the task while ON medication, and a subset (n = 9) completed both ON and OFF dopaminergic medication sessions.

Participants performed a peak-interval timing task with intermixed 3-second and 7-second trials. They were instructed to press a key when they estimated the target interval had elapsed. Visual distractions were included to discourage counting. Each participant completed 80 trials (40 per interval type). Only trials with a minimum of 20 valid keypresses per interval condition were included in analyses.

EEG was recorded using a 64-channel actiCAP system at 500 Hz. 

\begin{table}[!htbp]
\centering
\begin{tabular}{|c|c|}
\hline
Number of subjects & 120 (83 PD, 37 Controls) \\
\hline
Sessions per subject & 1–2 (PD), 1 (Control) \\
\hline
Average session length & 30.0 min \\
\hline
EEG channels & 64 \\
\hline
Sampling rate & 500 Hz \\
\hline
Total recordings & 129 \\
\hline
Hardware & actiCAP Brain Products \\
\hline
\end{tabular}
\smallskip
\caption{Summary of the Singh2021 dataset}
\end{table}

\subsection{TUAB}
TUAB is the second-largest annotated subset of the TUEG corpus \cite{TUHEEG}, collected at Temple University Hospital. It contains EEG recordings from 2,383 subjects that were classified as either ``normal'', if it fulfills certain characteristics, or as ``abnormal'', if it does not or if it contains patterns indicating pathological conditions \cite{abnormal}.

This very basic type of classification can serve as an important first step in deciding whether to inspect a given EEG recording closer and search for more concrete conditions.

Table \ref{tab:tuab} gives an overview of the statistics of the TUAB dataset.

\begin{table}[!htbp]
\centering
\begin{tabular}{|c|c|}
\hline
Number of subjects & 2,383 \\
\hline
Sessions per subject (average) & 1.3 \\
\hline
Average session length & 23 min \\
\hline
EEG channels & 19 \\
\hline
Sampling rates & $\{250, 256, 512\}$ Hz \\
\hline
Number of abnormal recordings & 1,472 \\
\hline
Number of normal recordings & 1,521 \\
\hline
\end{tabular}
\smallskip
\caption{Summary of the TUAB dataset}\label{tab:tuab}
\end{table}

\subsection{TUEP}
Epilepsy is a neurological disorder affecting an estimated 50 million people world-wide \cite{whoepilepsy}. In order to treat a patient with epilepsy, the condition must first be diagnosed, which can be achieved using EEG.

Like TUAB, TUEP \cite{tuep} is also an annotated subset of the TUEG corpus. It contains recordings from 100 patients with epilepsy and 100 patients without epilepsy. Despite the equal number of subjects, the number and total length of recordings from epileptic subjects exceeds that of non-epileptic subjects by a factor greater than 4, hence causing a significant class imbalance in the dataset.

A summary of basic statistics for TUEP are given in table \ref{tab:tuep}.

\begin{table}[!htbp]
\centering
\begin{tabular}{|c|c|}
\hline
Number of subjects & 200 \\
\hline
Sessions per subject (average) & 10.2 \\
\hline
Average session length & 18.6 min \\
\hline
EEG channels & 19 \\
\hline
Sampling rates & $\{250, 256, 400, 512, 1000\}$ Hz \\
\hline
Number of epilepsy recordings & 1,651 \\
\hline
Number of recordings without epilepsy & 390 \\
\hline
\end{tabular}
\smallskip
\caption{Summary of the TUEP dataset}\label{tab:tuep}
\end{table}

\subsection{TUAR}
The TUAR dataset \cite{tuar} contains annotations of ``artifacts'' in the EEG data. Artifacts are disturbances in an EEG recording that are not caused by the recorded brain, but rather by external factors. In TUAR, these include movements by the participant, such as chewing, shivers, agitation and eye movements, as well as electrode issues, like interferences or displacements.

As artifacts can be mistaken for other event types, like seizures \cite{tuar}, or can make the detection of an event with which they overlap more difficult, it is often desirable to detect and/or remove artifacts from a given EEG signal before attempting to detect patterns of interest. Hence, the aim of TUAR is aid the development of artifact detection techniques.

TUAR is a subset of the TUEG corpus, containing annotated EEG recordings from 213 patients, taken over 259 sessions. Table \ref{tab:tuar} presents some basic statistics of the dataset.

\begin{table}[!htbp]
\centering
\begin{tabular}{|c|r|r|}
\hline
Number of subjects & \multicolumn{2}{c|}{213} \\
\hline
Sessions per subject (average) & \multicolumn{2}{c|}{1.2} \\
\hline
Average session length & \multicolumn{2}{c|}{23 min} \\
\hline
EEG channels & \multicolumn{2}{c|}{19} \\
\hline
Sampling rates & \multicolumn{2}{c|}{$\{250, 256, 400, 512, 1000\}$ Hz} \\
\hline
Average event length & \multicolumn{2}{c|}{8 s} \\
\hline
\multirow{5}{*}{Number of events} & Eye Movement & 60,577 \\
\cline{2-3}
& Muscle Artifact & 81,623 \\
\cline{2-3}
& Electrode Artifact & 43,356 \\
\cline{2-3}
& Chewing & 7,741 \\
\cline{2-3}
& Shivers & 659 \\
\hline
\end{tabular}
\smallskip
\caption{Summary of the TUAR dataset}\label{tab:tuar}
\end{table}

\subsection{CHB-MIT}
CHB-MIT \cite{chbthesis} contains annotated seizures from the Children's Hospital Boston. In order to better understand how to counteract their seizures, the 23 subjects had their anti-seizure medication withdrawn for the duration of the study and their brain activity was subsequently recorded.
In total, 182 seizure events were annotated. It should be mentioned that the EEG channels are given as a bipolar montage. Hence, none of the foundation models was trained with any of CHB-MIT's channels. In order to still be able to train them on CHB-MIT, we decided to ignore the models' pre-defined channels and use those of CHB-MIT anyway, in an arbitrary (but fixed) order.

A summary of statistics of CHB-MIT can be found in table \ref{tab:chbmit}.
\begin{table}[!htbp]
\centering
\begin{tabular}{|c|c|}
\hline
Number of subjects & 23 \\
\hline
Sessions per subject (average) & 29.8 \\
\hline
Average session length & 1.4 h \\
\hline
EEG channels & 18 \\
\hline
Sampling rate & 256 Hz \\
\hline
Average seizure length & 62 s \\
\hline
Number of seizures & 182 \\
\hline
\end{tabular}
\smallskip
\caption{Summary of the CHB-MIT dataset}\label{tab:chbmit}
\end{table}

\subsection{Sleep-Telemetry}
Human sleep can be categorized into different phases, or ``stages''. Overall, one can distinguish rapid eye movement (REM) phases and non-REM phases, while the non-REM phases can be further differentiated by the “depth” of the sleep (i.e. how hard it is to wake someone up), into N1, N2, N3 and N4 phases (ordered by increasing depth). These phases can be well observed in an EEG recording and hence, this type of measure is used in many areas of the study of sleep.

In this paper, we use the publicly available Sleep-Telemetry \cite{sleep_telemetry} dataset, which was part of a study investigating the effects of temazepam medication on sleep behavior. It contains data from 22 subjects aged between 18 and 79 years. For each subject, two nights of nine hours were recorded, one night with temazepam intake, and one night with a placebo.
The dataset features three channels: Fpz-Cz, Pz-Oz and horizontal EOG (Electrooculogram). Similar to CHB-MIT, none of the foundation models was trained on any of these channels and hence, we again opt to ignore the model-channels.

An overview of some statistics of the Sleep-Telemetry dataset is given in table \ref{tab:sleeptelemetry}.

\begin{table}[!htbp]
\centering
\begin{tabular}{|c|r| r|}
\hline
Number of subjects & \multicolumn{2}{c|}{22} \\
\hline
Sessions per subject & \multicolumn{2}{c|}{2} \\
\hline
Average session length & \multicolumn{2}{c|}{8.6 h} \\
\hline
Channels & \multicolumn{2}{c|}{$2\times$ EEG, $1\times$ EOG} \\
\hline
Sampling rate & \multicolumn{2}{c|}{100 Hz} \\
\hline
Average event length & \multicolumn{2}{c|}{3.5 min} \\
\hline
\multirow{5}{*}{Number of events} & Wake Phase & 744 \\
\cline{2-3}
& N1 Phase & 1311 \\
\cline{2-3}
& N2 Phase & 1718 \\
\cline{2-3}
& N3 or N4 Phase & 1631 \\
\cline{2-3}
& REM Phase & 378 \\
\hline
\end{tabular}
\smallskip
\caption{Summary of the Sleep-Telemetry dataset}\label{tab:sleeptelemetry}
\end{table}

\section{Dataset Licensing and Usage Compliance}
\label{sec:license}
We reviewed and complied with the license or usage terms of all datasets included in EEG-Bench. Below is a dataset-by-dataset breakdown:

\begin{itemize}

    \item \textbf{TUAB}, \textbf{TUEP}, \textbf{TUAR} – Provided by the Neural Engineering Data Consortium (NEDC) under a public data use agreement. \\
    We have registered with NEDC and fully comply with their usage terms: (1) the dataset providers are acknowledged as requested, (2) we do not redistribute the data, (3) no attempts are made to re-identify subjects, (4) we use the data solely for research and non-malicious purposes, and (5) we agree to delete the data after use if required.

    \item \textbf{Cavanagh2018a}, \textbf{Cavanagh2018b}, \textbf{Cavanagh2019}, \textbf{Albrecht2019}, \textbf{Singh2018}, \textbf{Brown2020}, \textbf{Gruendler2009}, \textbf{Singh2020}, \textbf{Singh2021} – Licensed under \textit{Public Domain Dedication and License (PDDL) v1.0}. \\
    Used in accordance with public domain status for academic research.

    \item \textbf{CHB-MIT}, \textbf{Sleep-Telemetry} – Licensed under \textit{Open Data Commons Attribution License v1.0}. \\
    Used with appropriate attribution for research purposes.
    
\end{itemize}

All datasets are used strictly for non-commercial, academic purposes. No data is redistributed or altered in violation of its license, and no attempts at re-identification or deanonymization have been made.

\newpage

\end{document}